\def\year{2022}\relax
\documentclass[letterpaper]{article} % DO NOT CHANGE THIS
\usepackage{aaai22}  % DO NOT CHANGE THIS
\usepackage{times}  % DO NOT CHANGE THIS
\usepackage{helvet}  % DO NOT CHANGE THIS
\usepackage{courier}  % DO NOT CHANGE THIS
\usepackage[hyphens]{url}  % DO NOT CHANGE THIS
\usepackage{graphicx} % DO NOT CHANGE THIS
\usepackage{natbib}  % DO NOT CHANGE THIS AND DO NOT ADD ANY OPTIONS TO IT
\usepackage{caption} % DO NOT CHANGE THIS AND DO NOT ADD ANY OPTIONS TO IT

\DeclareCaptionStyle{ruled}{labelfont=normalfont,labelsep=colon,strut=off} % DO NOT CHANGE THIS
\usepackage{algorithm}
\usepackage{algorithmic}

\usepackage{newfloat}
\usepackage{listings}
\floatstyle{ruled}
\newfloat{listing}{tb}{lst}{}
\floatname{listing}{Listing}
\usepackage{subfigure}
\usepackage{array}
\usepackage{amsmath}
\usepackage{amssymb}
\usepackage{multirow}
\usepackage{diagbox}

\title{Improving Video-Text Retrieval by\\ Multi-Stream Corpus Alignment and Dual Softmax Loss}
\author{
    Xing Cheng\thanks{Interns at MMU, KuaiShou Inc.}, HeZheng Lin\footnotemark[1], XiangYu Wu \thanks{Corresponding author}, Fan Yang, Dong Shen
    \\
}
\affiliations{
    MMU \quad KuaiShou Inc.\\
    \{chengxing03,linhezheng,wuxiangyu,yangfan,shendong\}@kuaishou.com

}
\begin{document}

\maketitle

\begin{abstract}
Employing large-scale pre-trained model CLIP to conduct video-text retrieval task (VTR) has become a new trend, which exceeds previous VTR methods. Though, due to the heterogeneity of structures and contents between video and text, previous CLIP-based models are prone to overfitting in the training phase, resulting in relatively poor retrieval performance. In this paper, we propose a multi-stream Corpus Alignment network with single gate Mixture-of-Experts (CAMoE) and a novel Dual Softmax Loss (DSL) to solve the two heterogeneity. The CAMoE employs Mixture-of-Experts (MoE) to extract multi-perspective video representations, including action, entity, scene, etc., then align them with the corresponding part of the text. In this stage, we conduct massive explorations towards the feature extraction module and feature alignment module, and conclude an efficient VTR framework. DSL is proposed to avoid the one-way optimum-match which occurs in previous contrastive methods. Introducing the intrinsic prior of each pair in a batch, DSL serves as a reviser to correct the similarity matrix and achieves the dual optimal match. DSL is easy to implement with only one-line code but improves significantly. The results show that the proposed CAMoE and DSL are of strong efficiency, and each of them is capable of achieving State-of-The-Art (SOTA) individually on various benchmarks such as MSR-VTT, MSVD, and LSMDC. Further, with both of them, the performance is advanced to a great extent, surpassing the 
previous SOTA methods for around 4.6\% R@1 in MSR-VTT. 
%The code will be available after blind review.
The code will be available soon at \url{https://github.com/starmemda/CAMoE/}
\end{abstract}

\section{Introduction}
% \begin{figure}[t]
% \begin{center}
% \subfigure[single-stream and two-stream]{\includegraphics[scale=0.22]{one_and_two_stream.png}}
% \subfigure[multi-stream]{
% \includegraphics[scale=0.22]{multi-stream.pdf}\label{multi_stream_arc}}
% \end{center}
% \caption{A diagram of single-stream and two-stream architectures}
% \label{one_and_two_stream}
% \end{figure}

\begin{figure*}[h]
\begin{center}
\includegraphics[scale=0.19]{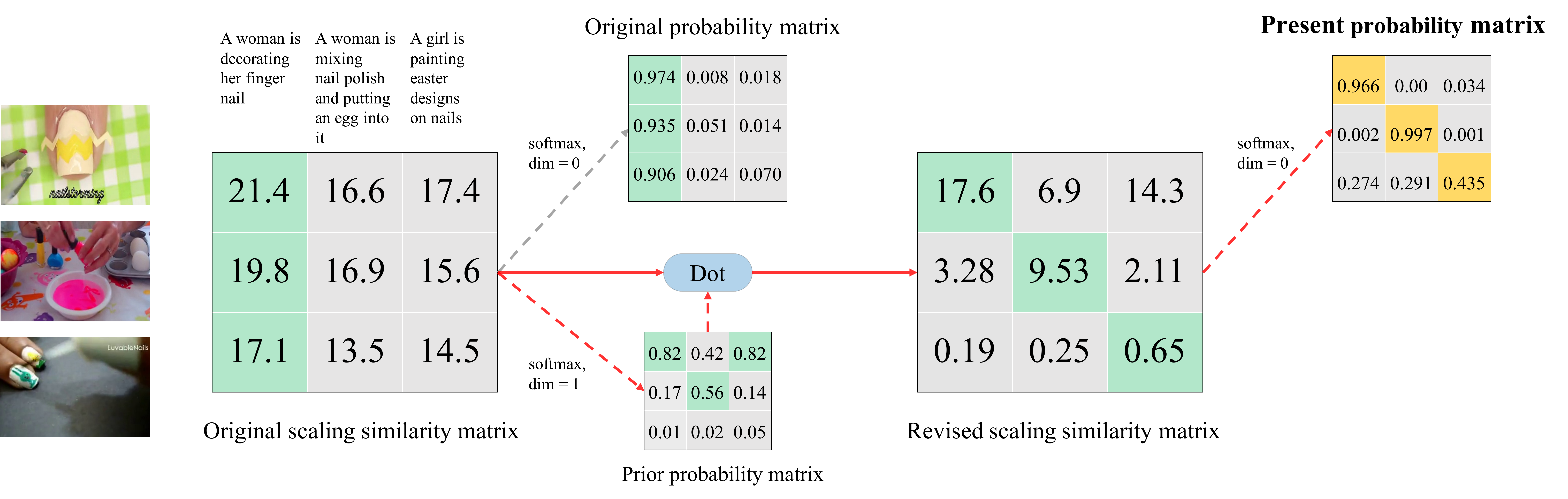}
\end{center}
\caption{A diagram of the heterogeneity of contents and Dual Softmax loss. The highlighted block denotes the maximum value in each row. The sentence "A woman is decorating her finger nail." describes broad content and can be paired with all videos painting on nails, so it is inferred with the maximum score for each row in the original similarity matrix. 
Considering the diagonal scores denote the ground truth and should be highlighted, a prior probability matrix is calculated in the cross direction. With the dot product of the prior and the original similarity, the diagonal part achieves the optimal.}
\label{confusion}
\end{figure*}

\subsection{Motivation}
The primary issue limiting VTR task presently is the heterogeneity between different modals, reflected in both structures and contents.

\textbf{The heterogeneity of structures.} This mainly 
lies in the impossibility of directly aligning the words in sentences with corresponding video frames \cite{jin2021hierarchical}. Single-stream or two-stream structures are applied to treat text and video as two independent parts for early or late fusion, which ignore the internal relevancy between frames and words, resulting in that the models require massive data to reach decent performance. In this paper, we assume that texts can be parsed into separate sentences with distinct aspects of information. Though directly aligning a word with a frame is unachievable, guiding the model to learn how to align cross modal information is possible. Referring to the example in Fig.\ref{framework}, the video is paired with the sentence "a boy is performing for an audience.", where "boy", "performing", "audience" are the keywords and can be categorized as "entity", "action", "entity" accordingly. We design several experts to learn corresponding  representations independently. In addition, a gating module is employed to measure their importance score and then strengthen the representation of the fusion expert. Such innovation brings little parameters and computations increment and surpassing the previous State-of-The-Art (SOTA) method on various benchmarks.

Previous work has taken a similar approach, either by simple part-of-speech tagging or by wielding multi-dimensional features on the video. HGR \cite{chen2020fine} and HCGC \cite{jin2021hierarchical} hypothesize that a text can be constructed into a hierarchical semantic graph structure, where lie sentence, action, entity embedding in the top, second, third level node, respectively. T2VLAD \cite{wang2021t2vlad} extracts features from the aspects of scene and action, and performs similarity matching with the representations of each local token and the global sentence, while HiT \cite{liu2021hit} conducts cross-matching between feature-level and semantic-level embedding. But they don't simultaneously decompose the video and text to conduct deep alignment, from where we proposed the multi-stream multi-task \cite{ruder2017overview} architecture, as shown in Fig.\ref{framework}
 
\textbf{The heterogeneity of contents and dual optimal-match hypothesis.} Another important contribution of this paper is the proposed problem that semantic and visual modals usually express in a different range of content. The example shown in Fig.\ref{confusion}, denotes the comparisons of the process calculating the final probability matrix for Video-to-Text retrieval. Although each video describes specific and explicit content, the corresponding text can be unspecific and fuzzy, which harms model training. The original method conducts the softmax for every single retrieval, ignoring the potential cross-retrieval information and leading to a confusing result. To solve this, we propose the dual optimal-match hypothesis based on the discovered phenomenon that when a Text-to-Video or Video-to-Text pair reaches the optimal match, the symmetric Video-to-Text or Text-to-Video score should be the highest. With this hypothesis, the corresponding 
DSL is designed to revise the predicted similarity score, significantly improving the performance. It introduces a prior probability matrix calculated in the cross direction to adjust the original scaling similarity matrix. By the dot product of the prior probability matrix and 
original scaling similarity matrix, we can filter the hard case with a high Video-to-Text similarity score but a low Text-to-Video similarity score. Referring to Fig.\ref{confusion}, the final probability matrix's maximum is adjusted to the diagonal, which indicates DSL's positive effect.

\subsection{Contributions}
Our main contributions can be concluded as:
\begin{itemize}
    \item We propose a visual-semantic data decomposing and multi-task constructing scheme for VTR task, and it is capable of being extended to other cross-modal tasks, such as image text generation, image text retrieval, image caption \cite{ding2021cogview,li2020oscar,zhang2021vinvl}. And some corresponding foundational explorations have been enforced.
    \item We state the problem of the contents heterogeneity in VTR for the first time and a corresponding dual optimal-match hypothesis is proposed, where massive undiscovered works can be done in the future to perfect the video text retrieval task. 
    \item The proposed CAMoE and Dual Softmax loss primarily specialized for contents heterogeneity advance the SOTA to a new level. We claim that CAMoE is a novel and promising architecture that can serve as the future cross-modal large-scale pre-training model. And DSL is one of the efficient ways to significantly improve the performance with eligible cost.
\end{itemize}

\section{Related work}

Video-text retrieval \cite{sivic2003video,yu2018joint,zhu2020actBert,lei2021less,gabeur2020multi,dzabraev2021mdmmt,mithun2018learning,arnab2021vivit}, as a task evolved from image-text retrieval \cite{faghri2017improving,frome2013devise,gong2014multi,gu2018look}, though developed for years, still largely follows the single-stream or two-stream architecture \cite{dzabraev2021mdmmt,zhu2020actBert,lei2021less,gabeur2020multi,fang2021clip2video,luo2021clip4clip}. For single-stream network, raw text and video frames are input into the network directly, and the cross-modal information is fused simultaneously. The two-stream network employs the separate text and video embedding extractors firstly and then matching the cross-modal embedding with specific fusion networks.

If dividing according to the ideas improving the performance, previous methods can be roughly inducted into two types: alignment-based and embedding-based methods.

The alignment-based method aims to decompose the video and text into a somehow regular structure to facilitate calculating the similarity. In the early practice, Le et al. \cite{le2016nii} uses bag-of-visual word model with geometric verification to search for shots With the query location. Markatopoulou et al. \cite{markatopoulou2017query} encode the decomposed query into related semantic concepts and then conduct concept matching with specific videos. Recently, HGR \cite{chen2020fine} is proposed to divide the sentence into three parts: events, action, and entity, with the hypothesis that all the captions own a relatively fixed hierarchical graph form. HCGC \cite{jin2021hierarchical} adopts the same sentence resolving strategy and introduces the hierar- chical cross-modal graph consistency learning into embedding space.

Unlike the alignment-based scheme, which is to a largely dependent on the regulation set by the designer, the methods of matching video and text features directly in the embedding space have begun to emerge with the development of big data and large models in recent years. MMT \cite{gabeur2020multi} puts forward the completed two-stream transformer framework solving video text retrieval for the first time. MDMMT \cite{dzabraev2021mdmmt} considers that the action word in the text is the key to encode effectively. Consequently, several excellent pre-training models pre-trained on various datasets are adopted to replace the original video encoder in MMT, and it concludes that CLIP performs best. CLIP4Clip \cite{luo2021clip4clip}, the previous SOTA, proposes three different similarity calculators to model temporal dependency between video frames. CAMoE reported in this paper reaches a consensus of adopting CLIP as pre-training parameters with MDMMT and CLIP4Clip.

\section{Method}
In this section, we will introduce the overall architecture of our approach and elaborate on the submodules within it.

\begin{figure*}[h]
\begin{center}
\includegraphics[scale=0.22]{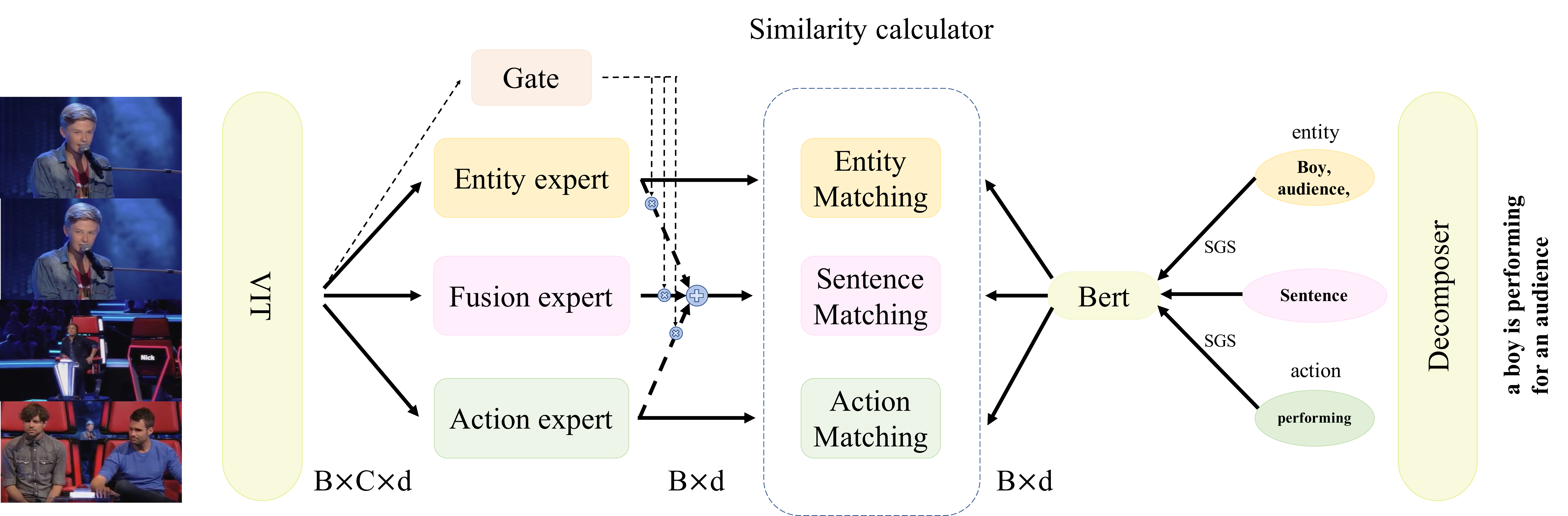}
\end{center}
\caption{An overview of the proposed CAMoE. SGS denotes the sentence generation strategy, which is explained in Fig.\ref{sgs_and_vfas}}
\label{framework}
\end{figure*}

\subsection{Overall Architecture}
In general, we convert the video and text into three-stream outputs according to the designed rules and conduct the consistency learning as Fig.\ref{framework}.

For the semantic side, the nouns and verbs of the text are selected out with the pre-trained part-of-speech tagging (POS) models \cite{toutanova2003feature,ratnaparkhi1996maximum}, and sequentially transformed into nouns sentence and verbs sentence by sentence generation strategy (SGS). Then we adopt Bert \cite{devlin2018Bert} pre-trained by CLIP \cite{radford2021learning} to encode 
them into semantic representations. Though we have tried to add a scene sentence to strengthen the background information, little texts can be extracted from scene parts. Too much irrelevant noise input may be detrimental for training.

Referring to the visual side, for fair comparison and keeping the efficiency, we adopt only Vit \cite{dosovitskiy2020image} pre-trained by CLIP \cite{radford2021learning} as the bottom feature extractor, which is the same as the previous SOTA method. Fusion, entity, and action experts are specially designed to learn distinct semantic matching from the bottom features. Note that the three tasks are separate but subordinative. A gate module \cite{ma2018modeling} is added to integrate entity and action representations with fusion ones, which improves fusion expert's performance.

After acquiring visual and semantic representations, they are matched in a similarity calculator, which leads to the loss value.

\subsection{Sentence generation strategy.}\label{section:SGS}

\begin{figure*}[t]
\begin{center}
\subfigure[The SGS of RKW]{\includegraphics[scale=0.25]{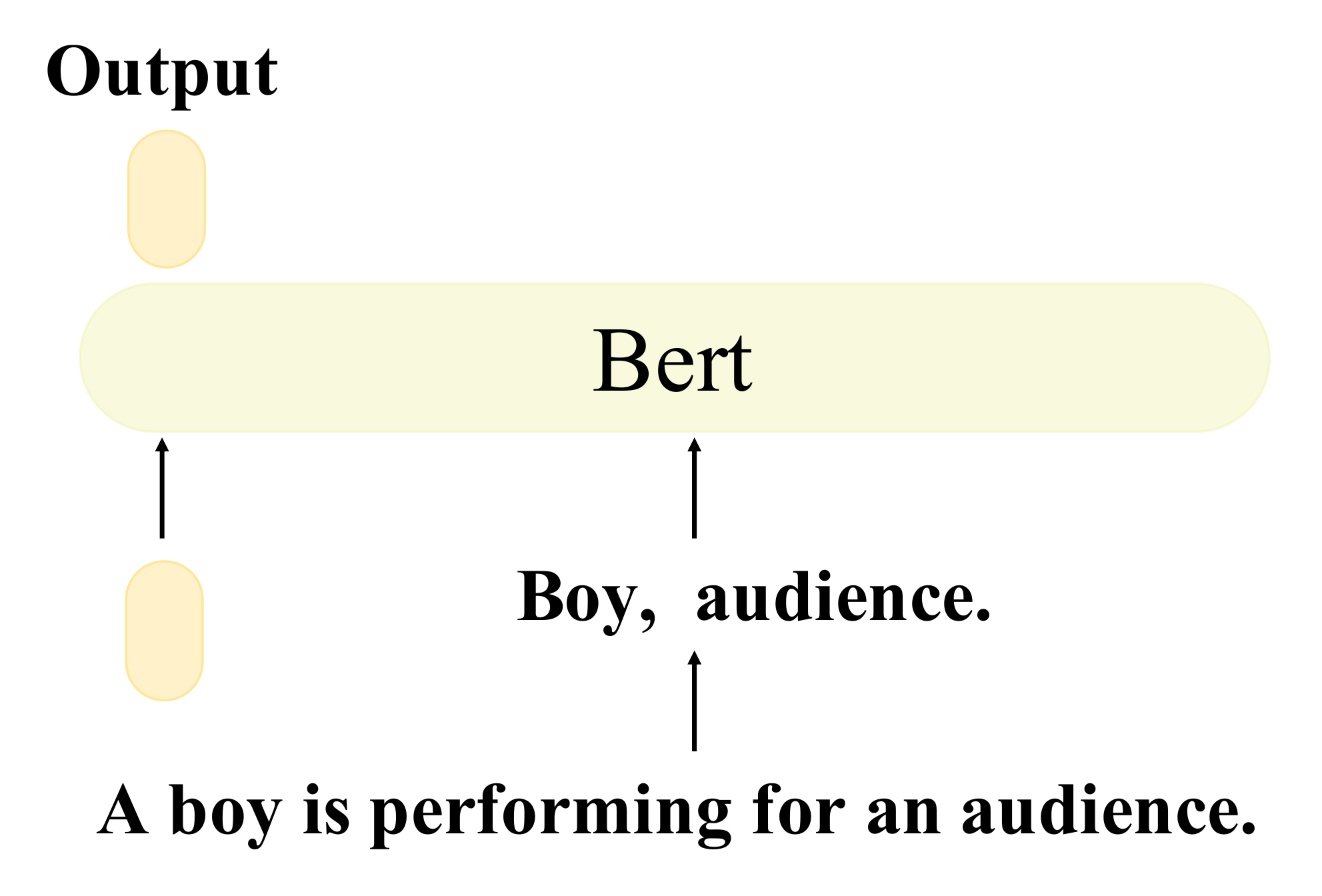}\label{RKW}}
\subfigure[The SGS of AKWE]{
\includegraphics[scale=0.25]{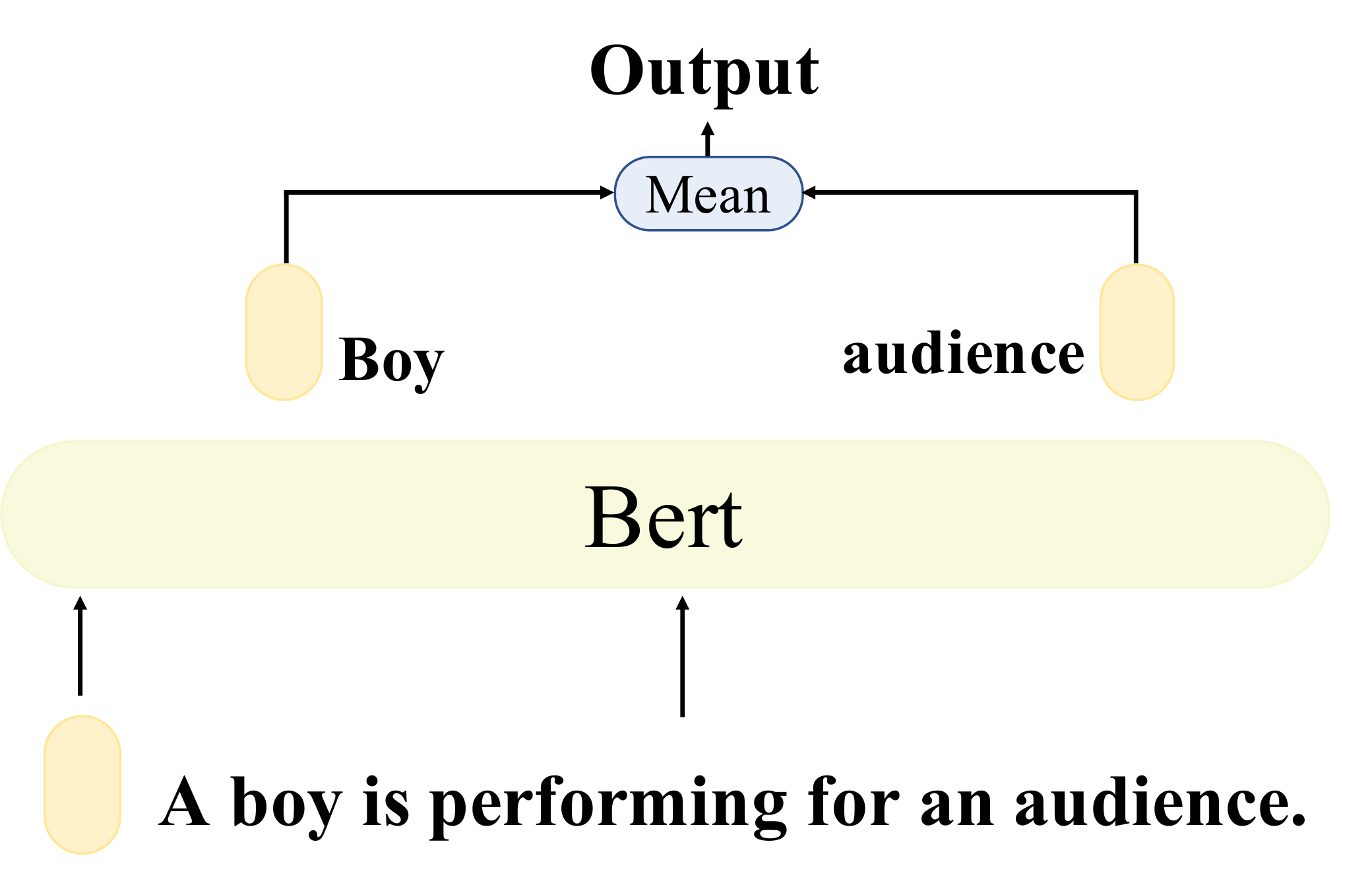}\label{AKWE}}
\subfigure[The SGS of MUW]{
\includegraphics[scale=0.25]{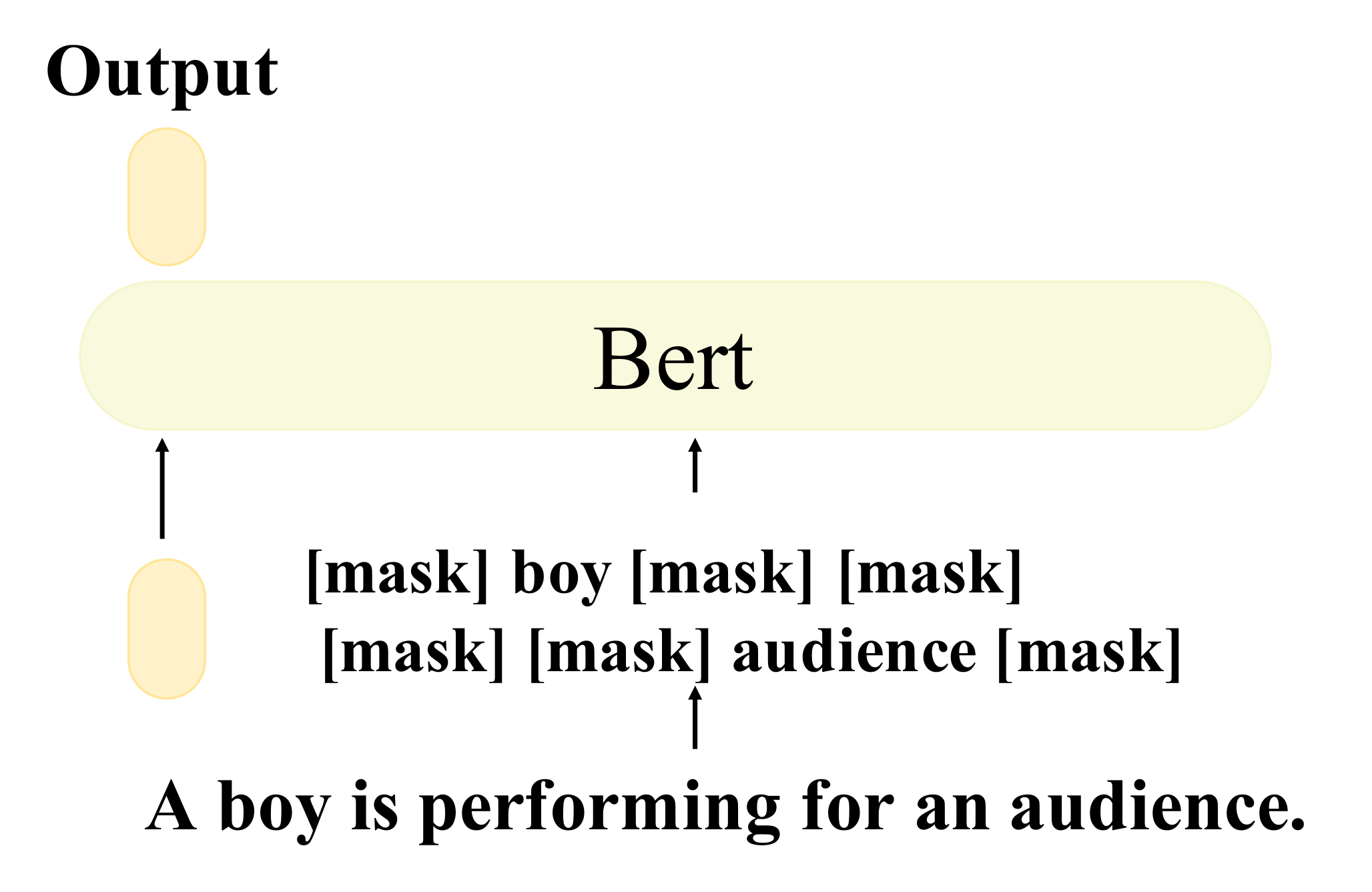}\label{MUW}}
\subfigure[The VFAS of mean pooling]{\includegraphics[scale=0.25]{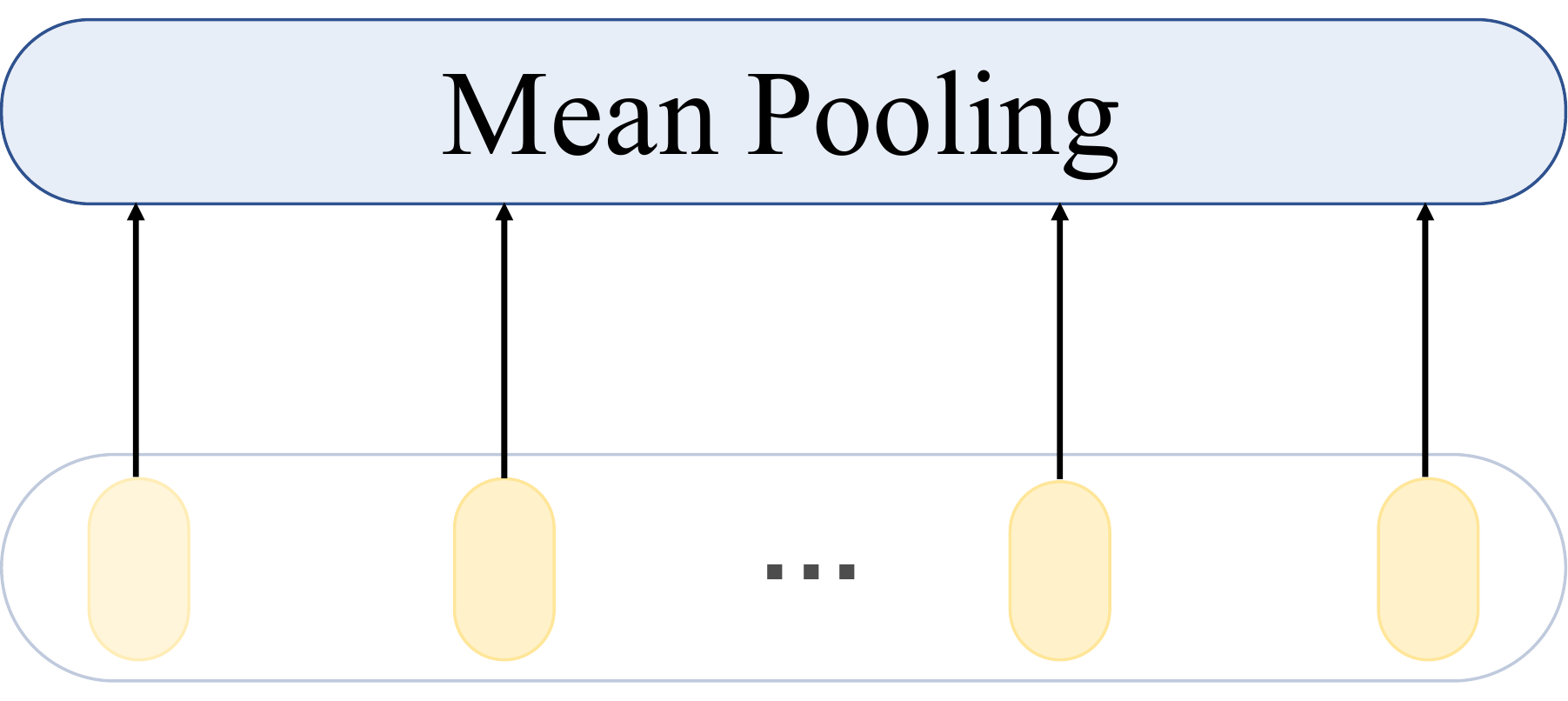}\label{mean_pool}}
\subfigure[The VFAS of se attention]{
\includegraphics[scale=0.25]{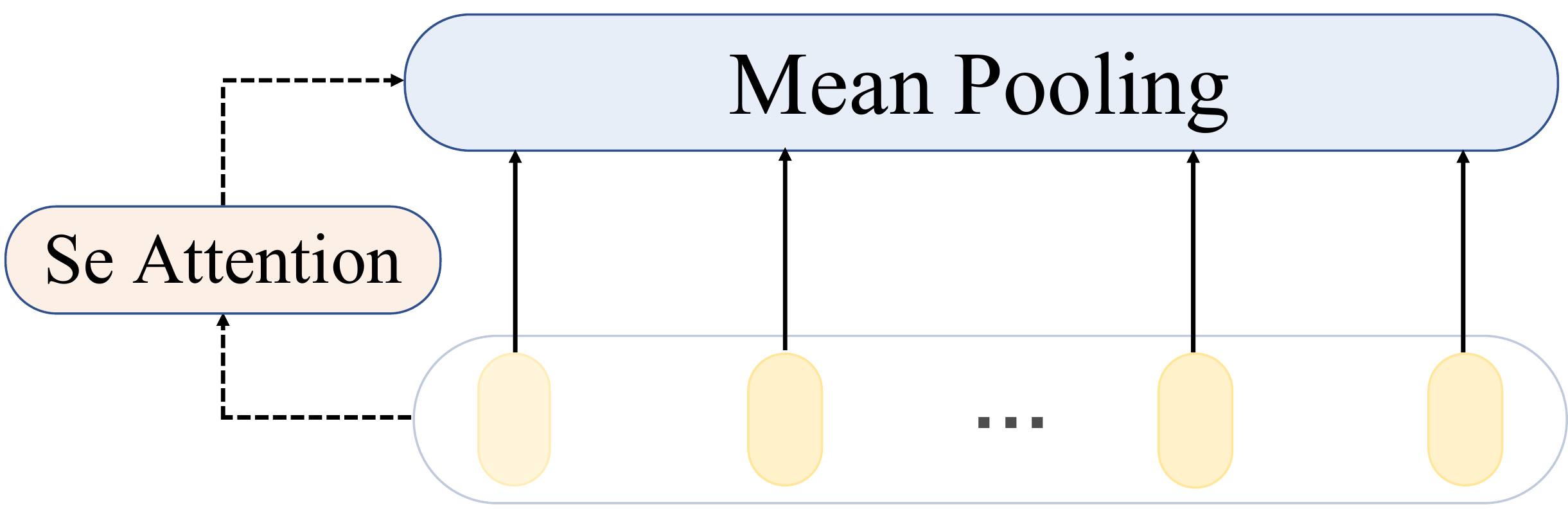}\label{se_mean_pool}}
\subfigure[The VFAS of self-attention]{
\includegraphics[scale=0.25]{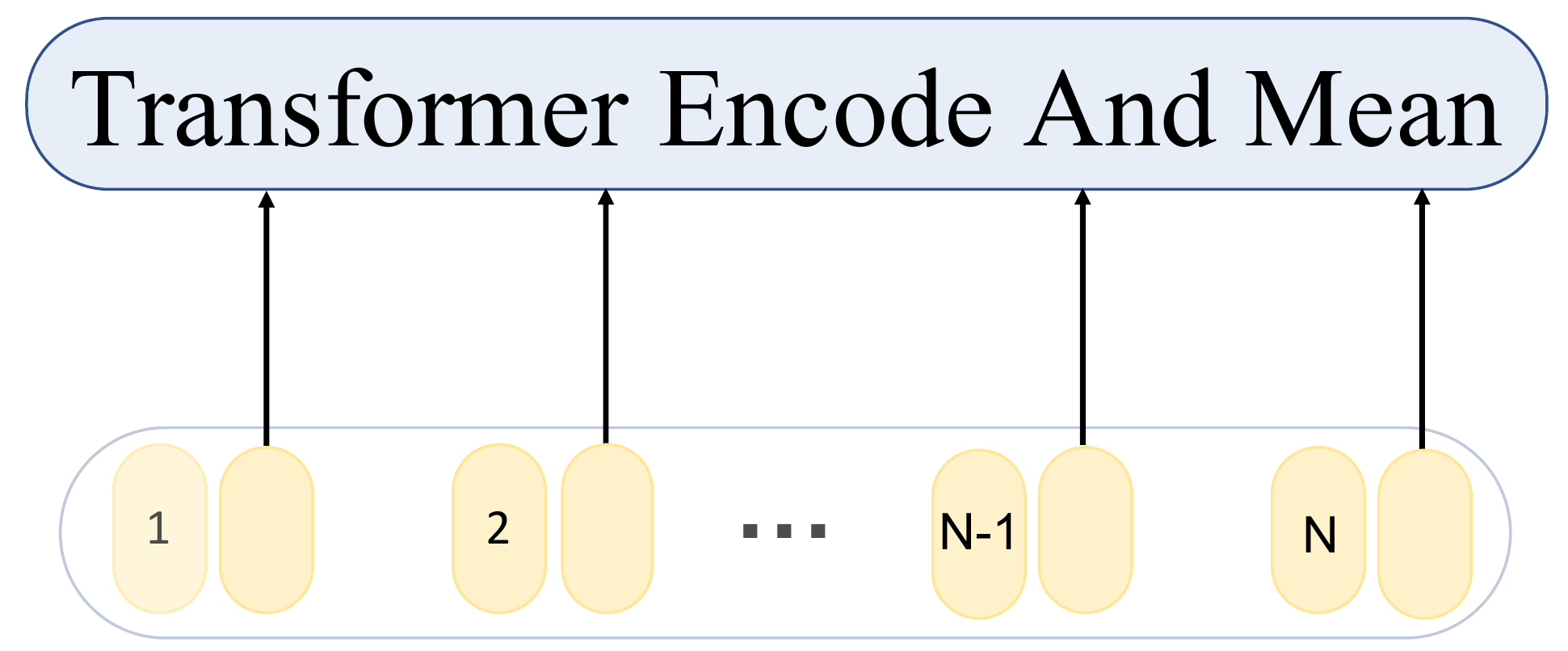}\label{transformer}}
\end{center}
\caption{The diagrams of three sentence generation strategies  (SGS) and three visual frames aggregation schemes (VFAS).}
\label{sgs_and_vfas}
\end{figure*}

As shown in Fig.\ref{sgs_and_vfas}, we test three different sentence generation strategies (SGS), named recombining keywords (RKW), averaging keywords embedding (AKWE), and masking unconsidered words (MUW), respectively. For RKW, the keywords are recombined into a sentence and then encoded by Bert, in which cls-token embedding represents the output. AKWE ignores no words, instead attentions the whole sentence, and adopts the average of keywords' output token embedding. To not change the sentence structure and 
keep networks from attention to all words, which will lead to overfitting, MUW takes another way. It masks all the non-keywords and then represents with cls-token, which is described as following:
\begin{eqnarray}
s_t = Bert(Mask(S))
\end{eqnarray}
Where $s_t$ denotes the semantic representations of task t, which is designed as sentence matching, entity matching, and action matching. $S$ denotes the input sentence.

\subsection{Visual Frames Aggregation Scheme.}
We uniformly sample $C$ frames for each video in batch $B$, and encode them into $x \in \mathbb{R}^{B\times{C}\times{d}}$ with a dimension $d$. We adopt three visual frames aggregation schemes and apply them to different experts or gates for various purposes, as Fig.\ref{sgs_and_vfas}. 
\begin{itemize}
    \item \textbf{Mean pooling} averages all frame features from a video directly:
        \begin{eqnarray}
            v = \sum_{i}^{C} x_i
        \end{eqnarray}
        Where $v$ denotes the aggregation module output.
    \item \textbf{Squeeze-and-Excitation attention (se attention)}\cite{hu2018senet} conducts average pooling firstly, and then calculate the importance score of each frame following feed forward networks(FFN) and sigmoid function:
        \begin{eqnarray}
            &scores = Sigmoid(FFN(AV(x)))\\
            &v = scores \cdot x
        \end{eqnarray}
        Where $scores, Sigmoid, FFN$, and $AV$ denote importance scores, sigmoid function, FFN, and average pooling operation.
    \item \textbf{Self-attention} \cite{vaswani2017attention} projects each frame feature into key $K$, query $Q$, value $V$ and then utilizes the matching degree of $K$ and $Q$ as the projection scores of $V$.
        \begin{eqnarray}
            &K, Q, V = FFN(x+p)\\
            &v = FFN(Softmax(\frac{QK^T}{\sqrt{d_K}})V)
        \end{eqnarray}
    Where p denotes the position embedding, and $d_K$ represents the dimension of $K$
\end{itemize}

\subsection{Experts and Gating network.}
We employ different frame aggregation schemes for foreign experts and gates. Specifically, when the gating network and fusion expert adopt se attention, entity expert, and action expert adopt self-attention, the proposed CAMoE performs best.
Relevant explorations are exhibited in Ablation Studies.

Since there are somehow subordinate relationships among the three tasks, we find the output of fusion expert $v_F$ will result in the overfitting of entity matching and action matching when the gate is added to these two tasks. This may be because that fusion expert is paired with the whole sentences, such abundant information will make the other learning tasks too simple. So only fusion expert adopts a gating module:

\begin{eqnarray}u
 &v_F^g = \sum_{i=1}^{3} g(x)_i e_i(x)\\
 &v_E^g = v_E\\
 &v_A^g = v_A
\end{eqnarray}
where $x$ denotes the input of mixture-of-gate, $g$ and $e$ mean gate and expert network. $v_F^g, v_E^g,$ and $v_A^g$ represent the visual input of the fusion, entity, and action matching unit.
The gate network is composed of aggregation networks, a single layer perceptron, and a softmax layer calculating the importance scores among experts:
\begin{eqnarray}
 g(x) = softmax(W^p \times AGG(x))
\end{eqnarray}
where $W^p \in \mathbb{R}^{d\times{E}}$ represents the projection matrix, and $AGG$ denotes the aggregation networks.

\subsection{Loss function.}
\begin{figure}[h]
\begin{center}
\includegraphics[scale=0.12]{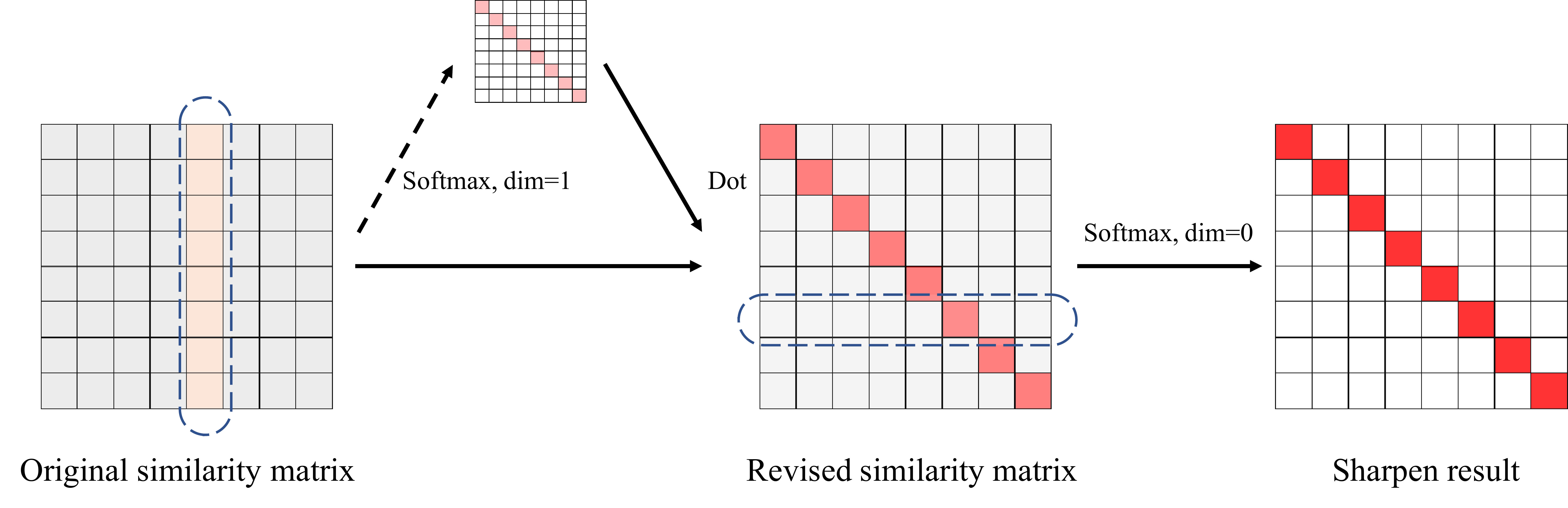}
\end{center}
\caption{The diagram of DSL.}
\label{cross_softmax}
\end{figure}

The proposed Dual Softmax loss is based on symmetric cross-entropy loss. Every text and video are calculated the similarity with other videos or texts, which should be maximum in terms of the ground truth pair. The original symmetric cross-entropy loss is as below:
\begin{eqnarray}
 & L_{t}^{v2t} = -\frac{1}{B}\sum_{i}^{B}{\rm log} \frac{ exp(l \cdot sim(v_i,s_i))}{\sum_{j=1}^B exp(l \cdot sim(v_i,s_j))} \\
 & L_{t}^{t2v} = -\frac{1}{B}\sum_{i}^{B}{\rm log} \frac{exp(l \cdot sim(v_i,s_i))}{\sum_{j=1}^B exp(l \cdot sim(v_j,s_i))} \\
 & L = \sum_{t}L_{t}^{v2t}+L_{t}^{t2v}
\end{eqnarray}
Where $t$ denotes sentence matching, entity matching, action matching, respectively. $i$ and $j$ denote the sample index in the batch. $l$ denotes a logit scaling parameter. We adopt a uniform one for all experiments in this paper, which is the same as CLIP. $sim$ represents the cosine similarity function:
\begin{eqnarray}
 & sim(v_i,s_i) = \frac{v_i \cdot s_i}{||v_i|| \cdot ||s_i||}
\end{eqnarray}
As for the DSL, a prior are introduced to revise the similarity score:
\begin{eqnarray}
 & L_{t}^{v2t} = -\frac{1}{B}\sum_{i}^{B}{\rm log} \frac{ exp(l \cdot sim(v_i,s_i) \cdot Pr_{i,i}^{v2t})}{\sum_{j=1}^B exp(l \cdot sim(v_i,s_j) \cdot Pr_{i,j}^{v2t})} \\
 & L_{t}^{t2v} = -\frac{1}{B}\sum_{i}^{B}{\rm log} \frac{exp(l \cdot sim(v_i,s_i) \cdot Pr_{i,i}^{t2v})}{\sum_{j=1}^B exp(l \cdot sim(v_j,s_i) \cdot Pr_{j,i}^{t2v})}
\end{eqnarray}
Where $Pr^{v2t}$, $Pr^{t2v}$ denotes the prior matrix for Video-to-Text and Text-to-Video task, $temp$ represents a temperature hyper-parameter to smooth the gradients. 
\begin{eqnarray}
 & Pr_{i,j}^{v2t} = \frac{exp(temp \cdot sim(v_i,s_i))}{\sum_{j=1}^B exp(temp \cdot sim(v_j,s_i))} \\
 & Pr_{i,j}^{t2v} = \frac{ exp(temp \cdot sim(v_i,s_i))}{\sum_{j=1}^B exp(temp \cdot sim(v_i,s_j))}
\end{eqnarray}
The specific process is as the diagram shown in Fig.\ref{cross_softmax}

It is worth noting that the practical difference lies in the introduced prior calculated in the cross direction. Multiplying the prior with the original similarity matrix imposes an efficient constraint and can help to filter those single side match pairs. As a result, DSL highlights the one with both great Text-to-Video and Video-to-Text probability, conducting a more convincing result.

\section{Experiments}
\subsection{Experimental Settings}
\subsubsection{Datasets}
We conduct the experiments on the benchmarks of MSR-VTT\cite{xu2016msr}, MSVD\cite{chen2011collecting}, and LSMDC\cite{rohrbach2015long}.

\begin{itemize}
    \item \textbf{MSR-VTT}
We employ the MSR-VTT\cite{xu2016msr} as the primary dataset, the most frequently researched object in video text retrieval. This dataset consists of 10000 videos, each 10 to 32s in length and 20 items in cation. To be precise, not all the cations are paired with the whole content of the corresponding video. Some may describe only a short clip, which makes the task more difficult. We report the result on 1K-A split\cite{gabeur2020multi}, in which 9k videos and 18w captions are used to train and another 1k videos as the test set. 
    \item \textbf{MSVD}\cite{chen2011collecting} is composed of 1970 videos with a split of 1200, 100, and 670 as the train, validation, and test set, respectively. Each video is paired with approximate 40 captions and ranges from 1 to 62 seconds.
    \item \textbf{LSMDC}\cite{rohrbach2015long,rohrbach2017movie} contains 118081 videos and equal captions extracted from 202 movies with a split of 109673, 7408, and 1000 as the train, validation, and test set. Every video is selected from movies ranging from 2 to 30 seconds.
\end{itemize}

\subsection{Metric.}
We conform to the standard metric settings as \cite{luo2021clip4clip}, which reports Recall at rank K (R@K), median rank (MdR) and mean rank (MnR). Generally, the higher R@K and lower MdR, MnR signify better performance.

\subsection{Implementation Details.}
We uniformly sample 16 frames for each video. The dimension of visual and semantic embedding is 512. Bert, Vit (ViT-B/32), and logit scaling $l$ are the same with CLIP. The learning rate of Bert and Vit are set to be 1e-7, and other parameters' are 1e-4. The optimizer and scheduler are Adam\cite{kingma2014adam} and warmup\cite{goyal2017accurate}. The sentence generation strategy adopts MUW. Visual frames aggregation scheme takes se attention for gate and fusion expert, self-attention for entity expert and action expert.

\subsection{Experimental Results}
\begin{itemize}
    \item \textbf{MSR-VTT}
    
\begin{table*}[h]
\centering
\caption{Experimental results of comparison with previous excellent methods on MSR-VTT dataset.}
\label{msrvtt_result}
\setlength{\tabcolsep}{1mm}{
\begin{tabular}{c|c|lllll|lllll}
\hline
\multicolumn{2}{c}{\qquad }&
\multicolumn{5}{c}{Text-to-Video Retrieval} &\multicolumn{5}{c}{Video-to-Text Retrieval}
\\
\hline
 
 \ &Model & R@1 & R@5 & R@10 &MdR &MnR & R@1 & R@5 & R@10 &MdR &MnR\\
\hline
\multirow{3}*{Others} & Collaborative Experts\cite{Liu2019a} &20.9 &48.8 &62.4 &6.0 &28.2 &- &- &- &- &- \\

&MMT\cite{gabeur2020multi} &24.6 &54.0 &67.1 &4.0 &26.7 &- &- &- &- &- \\

& FROZEN\cite{bain2021frozen} &31.0 &59.5 &70.5 &3.0 &- &- &- &- &- &- \\
\hline

\multirow{4}*{CLIP-based} &
CLIP\cite{radford2021learning} &31.2 &53.7 &64.2 &4.0 &- &27.2 &51.7 &62.6 &5.0 &- \\

 & MDMMT\cite{dzabraev2021mdmmt} &38.9 &69.0 &79.9 &2.0 &16.5 &- &- &- &- &-\\

& CLIP4Clip\cite{luo2021clip4clip} & 44.5 &71.4 &81.6 &2.0 &15.3 &42.7 &70.9 &80.6 &2.0 & - \\

%& CLIP2Video\cite{fang2021clip2video} & 45.6 & 72.6 & 81.7 & 2.0 &14.6 & 43.3 &72.3 & 82.1 &2.0 &10.2\\
\hline
& CAMoE(ours) & 44.6 & 72.6 & 81.8 & 2.0 &13.3 & 45.1 &72.4 & 83.1 &2.0 &10.0\\
& CAMoE+DSL(ours) &\textbf{47.3} & \textbf{74.2} &\textbf{84.5} &\textbf{2.0} &\textbf{11.9} & \textbf{49.1} &\textbf{74.3} &\textbf{84.3} &\textbf{2.0} &\textbf{9.9}\\

\hline
\end{tabular}}
\end{table*}

Referring to Table.\ref{msrvtt_result}, firstly, we can conclude that the models based on CLIP usually perform better than others. It isn't surprising due to its powerful generalization ability after learning from massive data. The proposed CAMoE automatically parses the original single task into multi-tasks and surpasses previous SOTA in various metrics with a standard loss function. If adopting the Dual Softmax loss, it achieves a higher SOTA. The R@1 is of approximate 2.8\% and 6.4 \% increments, respectively. It's interesting that Dual Softmax loss imposes a more important effect on Video-to-Text than on Text-to-Video, which is consistent with our hypothesis that texts' descriptions can be unspecific and be matched with several videos. As shown in Fig.1 (in Appendix), the advantages of CAMoE mainly reflect on alleviating overfitting. The main reason is that the extra two match units provide more complicated but critical targets and prevent the model from falling into local optimum by learning unprofitable words.

    \item \textbf{MSVD}

On the test dataset of MSVD as shown in Fig.\ref{msvd_result}, the purely model-optimized CAMoE increases the Text-to-Video R@1 SOTA to 46.9 and lowers the prediction mean rank by 0.2\%. And Dual Softmax plays a more significant role by improving R@1 by 2.9\% over previous SOTA.

\begin{table}[h]
\centering
\caption{Experimental results of comparison with previous excellent methods on MSVD dataset. All metrics are measured for Text-to-Video Retrieval.}
\label{msvd_result}
\begin{tabular}{c|c|c|c|c}
\hline
Model &R@1  &R@5 &R@10 &MnR \\
\hline
Collaborative Expert &19.9 &49.0 &63.8 &23.1 \\
FROZEN &33.7 &64.7 &76.3 &- \\
CLIP &37 &64.1 &73.8 &- \\
CLIP4Clip &46.2 &76.1 &84.6 &10.0 \\
\hline
CAMoE &46.9 &76.1 &85.5 &9.8 \\
CAMoE+DSL &\textbf{49.8} &\textbf{79.2} &\textbf{87.0} &\textbf{9.4} \\
\hline
\end{tabular}
\end{table}

    \item \textbf{LSMDC}

The result on LSMDC is shown in Fig.\ref{lsmdc_result}. LSMDC differs from other datasets adopted in this paper, where it contains the most videos, and each video pairs with only one caption. So present neural networks generally perform poorly, while CAMoE still improves 0.9\% in R@1. With Dual Softmax loss, the improvement turns to be 4.3\%

\begin{table}[h]
\centering
\caption{Experimental results of comparison with previous excellent methods on LSMDC dataset. All metrics are measured for Text-to-Video Retrieval.}
\label{lsmdc_result}
\begin{tabular}{c|c|c|c|c}
\hline
Model &R@1  &R@5 &R@10 &MnR \\
\hline
CLIP &11.3 &22.7 &29.2 &- \\
FROZEN &15.0 &30.8 &39.8 &- \\
MDMMT &18.8 &38.5 &47.9 &58.0 \\
CLIP4Clip &21.6 &41.8 &49.8 &58.0 \\
\hline
CAMoE &22.5 &42.6 &50.9 &56.5 \\
CAMoE+DSL &\textbf{25.9} &\textbf{46.1} &\textbf{53.7} &\textbf{54.4} \\
\hline
\end{tabular}
\end{table}

\end{itemize}

\section{Ablation Studies}

\subsection{Structure Design}
Since the CAMoE architecture belongs to multi-tasks with different sentence input, we compare it with the single task and multi-tasks inputting the same captions. Moreover, gates added to all tasks are also explored. As shown in Table.\ref{ab_sd}, multi-task with distinct input and single gate outperforms others, which indicates the efficiency of experts learning information from particular aspects. Combining the gate with all experts improves little, this may be because that fusion expert, which learns much better, will lead the other experts struck in local optimum through gating module. So we do not recommend introducing gates to all tasks that own subordinate relationships.

\begin{table}[h]
\centering
\caption{Ablation study for the architecture design on MSR-VTT. MTAC denotes multi-task with all captions input.}
\label{ab_sd}
\begin{tabular}{c|c|c|c|c}
\hline
&\multicolumn{2}{c|}{Text-to-Video} & \multicolumn{2}{c}{Video-to-Text} \\
\hline
method &R@1  &R@5 &R@1 &R@5 \\
\hline
single task &43.1 &70.8 &41.8 &70.2 \\
MTAC &43.8 &71.5 &42.6 &71.3 \\
multi-gate &43.5 &71.0 &43.3 &71.6 \\
\hline
CAMoE &\textbf{44.6} &\textbf{72.6} &\textbf{45.1} &\textbf{72.3} \\
\hline
\end{tabular}
\end{table}

\subsection{Sentence Generation Strategy and Visual Frames Aggregation Scheme}

%The relevant explorations are demonstrated in Table.\ref{ab_sgs}. 

\begin{table}[h]
\centering
\caption{Ablation study for the sentence generation strategy and visual frames aggregation scheme on MSR-VTT. RKW and AKWE are described in Fig.\ref{sgs_and_vfas}. mean pooling, se attention, self-attention denote adopting the scheme for all gates and experts. CAMoE
employs MUW for sentence generation, se attention for gate and fusion experts, self-attention for entity and action experts.}
\label{ab_sgs}
\begin{tabular}{c|c|c|c|c}
\hline
&\multicolumn{2}{c|}{Text-to-Video} & \multicolumn{2}{c}{Video-to-Text} \\
\hline
method &R@1  &R@5 &R@1 &R@5 \\
\hline
RKW &43.1 &71.1 &44.8 &72.2 \\
AKWE &43.1 &70.9 &43.1 &71.2 \\
\hline 
mean pooling &42.7 &70.9 &44.0 &71.7 \\
se attention &43.5 &72.3 &44.6 &72.1 \\
self-attention &39.2 &69.7 &39.0 &69.4 \\
\hline
CAMoE (MUW) &\textbf{44.6} &\textbf{72.6} &\textbf{45.1} &\textbf{72.3} \\
\hline
\end{tabular}
\end{table}

The experimental results in Table.\ref{ab_sgs} are in line with our hypothesis. RKW destroys the original sentence organization, AKWE inputs the whole sentence and results in overfitting, which is contrary to our of intention making professional expert learn specific features.

As for the frames integration methods, although there is consensus that the capacity of self-attention, which is exactly appropriate for entity and action experts who urge for complicated embedding space transformation, is more advanced than that of mean pooling and se attention, it does not always perform best on limited dataset due to the additionally introduced parameters and computational effort. Se attention only increases a very small number of parameters, allowing the model to automatically learn to attention to the keyframes in each video. Finally, we conclude that when the gate and fusion experts adopt the se attention, entity and action experts employ the self-attention, the proposed method performs best. 

\subsection{Loss optimization}

Referring to the result in Table.\ref{msrvtt_result}, Table.\ref{msvd_result}, and Table.\ref{lsmdc_result}, Dual Softmax loss makes excellent progress in all metrics for all benchmarks. To further prove the generalization of Dual Softmax, we also test it on other methods such as CLIP, FROZEN, and CLIP4Clip. Referring to Table.\ref{ab_method}, Dual Softmax provides a fantastic improvement for all models, and even more significant increments with lower based recall scores. For the example of CLIP V2T-R@1, around 10 points of gain is given, but as for CLIP4Clip, whose original R@1 exceeds 40\%, the increase drops down to 4.9 \%. For a method with weak generalization, the local optimum caused by content heterogeneity may be one of the most significant reasons. And we infer that it is still a valuable problem to explore for future works.

\begin{table}[h]
\centering
\caption{Ablation study of Dual Softmax for various methods. T2V and V2T denote Text-to-Video and Video-to-Text.}
\label{ab_method}
\setlength{\tabcolsep}{0.3mm}{
\begin{tabular}{c|cc|cc}
\hline

\diagbox{Method}{Loss}
&\multicolumn{2}{c|}{Original loss} & \multicolumn{2}{c}{DSL} \\
\hline
\  &T2V-R@1  &V2T-R@1 &T2V-R@1 &V2T-R@1 \\
\hline
CLIP &31.2 &27.2 &35.6 &37.2 \\
FROZEN &31.0 &- &45.5 &- \\
CLIP4Clip &44.5 &42.7 &47.0 &47.6 \\
\hline 
\end{tabular}}
\end{table}

\section{Quantitative Analysis and Visualization}

\subsection{Expert Importance Analysis}
It will be interesting to figure out how much each experts' information occupies the results and the accuracy each expert can reach alone. So we test the metrics from embedding space produced by each expert. Referring to Table.\ref{seperate_test}, the fusion expert matches with the whole sentence and performs best. Entity expert reaches about 27\%, while action expert reaches 8.4\% for T2V and 4.2\% for V2T, which is the worst. In addition, the average weights of the three experts calculated by the gate on the MSR-VTT test dataset are 0.63, 0.29, and 0.08, which indicates the two extra experts make an impact. We infer the reason for the action expert's lower accuracy and weight is that the CLIP we adopt is pre-trained on the image-text pair dataset, which is more equipped with entity information. So, the future solutions that can largely improve this task may rely on large-scale video-text pre-training or replace the uniform feature extractor of CAMoE architecture with the professional entity, action extractors. This indicates that CAMoE is one of the most excellent solutions with little computation increments and can serve as a pretrain architecture for future work.

\begin{table}[h]
\centering
\caption{Separate test for each Expert.}
\label{seperate_test}
\setlength{\tabcolsep}{0.5mm}{
\begin{tabular}{c|cc|cc}
\hline

\diagbox{Expert}{Loss}
&\multicolumn{2}{c|}{Original loss} & \multicolumn{2}{c}{DSL} \\
\hline
\  &T2V-R@1  &V2T-R@1 &T2V-R@1 &V2T-R@1 \\
\hline
Entity &27.7 &26.9 &31.1(+3.4) &30.5(+3.6) \\
Action &8.4 &4.2 &8.9(+0.5) &4.9(+0.7) \\
Fusion &44.6 &45.1 &47.3(+2.7) &49.1(+4.0) \\
\hline
\end{tabular}}
\end{table}

\subsection{Dual Softmax Visualization}
\begin{figure}[h]
\begin{center}
\includegraphics[scale=0.22]{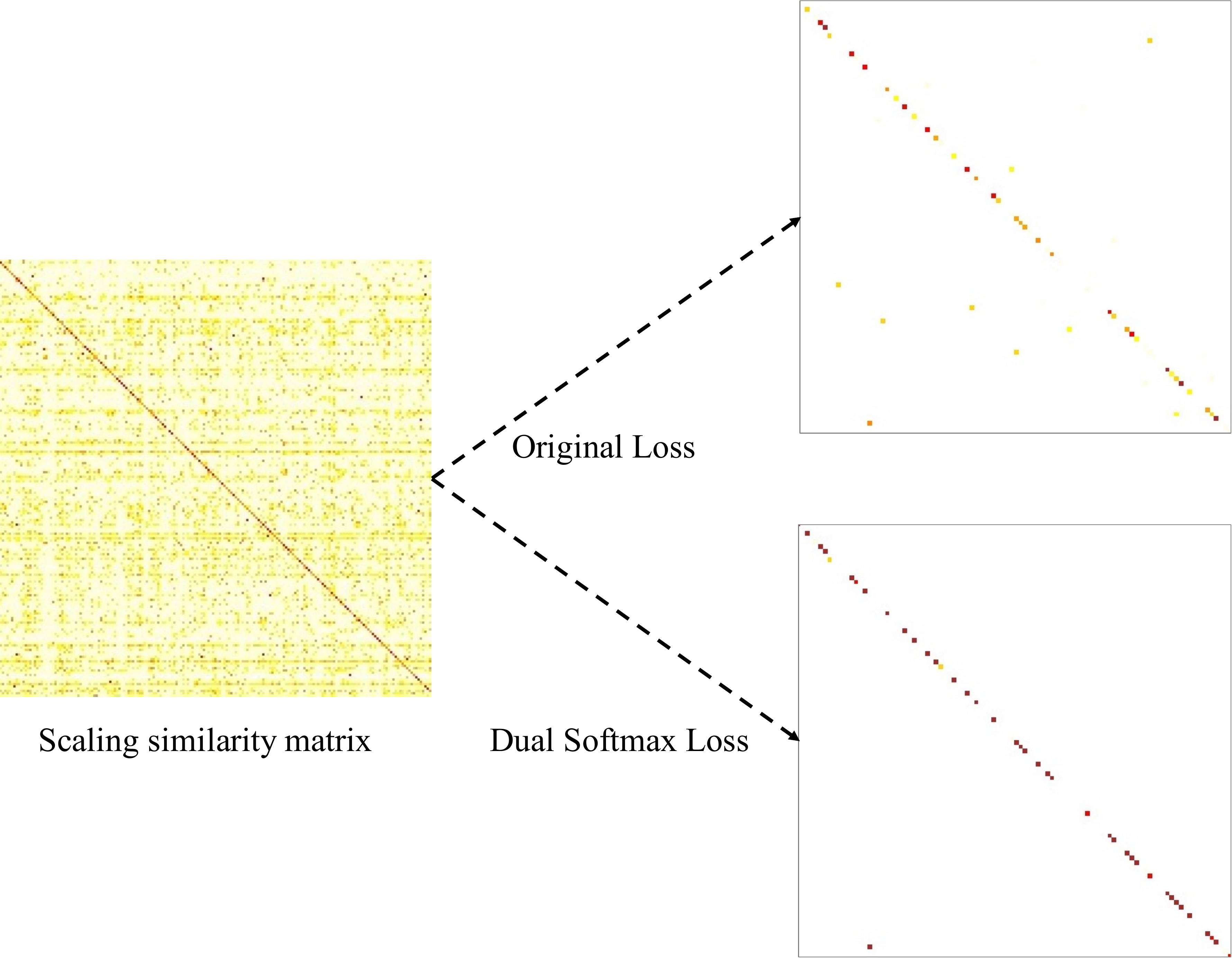}
\end{center}
\caption{The comparison of original inference method and Dual Softmax.}
\label{visualization}
\end{figure}

To vividly illustrate the role of Dual Softmax, we compare the inferred probability matrix of Dual Softmax with that of the previous one as Fig.\ref{visualization}. The visualization indicates that Dual Softmax improves in two aspects: 
\begin{itemize}
    \item Filtering the outliers.
    \item Sharpening the crucial and confidence points.
\end{itemize}
We reason that the constraints of cross direction correct partial borderline scores by introducing a prior probability matrix. For instance, when calculating the Video-to-Text matrix, a video can be paired with multiple unspecific and generalized texts. If introducing the prior Text-to-Video probability matrix, the sample with a high Video-to-Text similarity score but low Text-to-Video probability will be ignored, which filters the outliers and then leads to sharpening the 
convincing points.

\section{Conclusion}
In this paper, we identify the data heterogeneity of structure and content in the field of video-text retrieval. For the structure heterogeneity, a multi-stream corpus alignment architecture is proposed and reaches SOTA. We suggest the future work focus on extending CAMoE's uniform feature extractor to professional extractors or adopting CAMoE as large-scale pre-training architecture.
To solve the error caused by the confusing sentence that may match with over one video, Dual Softmax loss is proposed based on the dual optimal-match hypothesis and surprisingly achieved significant improvement with little extra training burden, which indicates its wide range of application scenarios in the industry and academia.

% Use \bibliography{yourbibfile} instead or the References section will not appear in your paper

%\bibliographystyle{unsrt}
\bibliography{myref}

\appendix
\section{The Demonstration of Preventing Overfitting}
\begin{figure}[t]
\begin{center}
\subfigure[The loss-epoch diagram of the single task]{\includegraphics[scale=0.45]{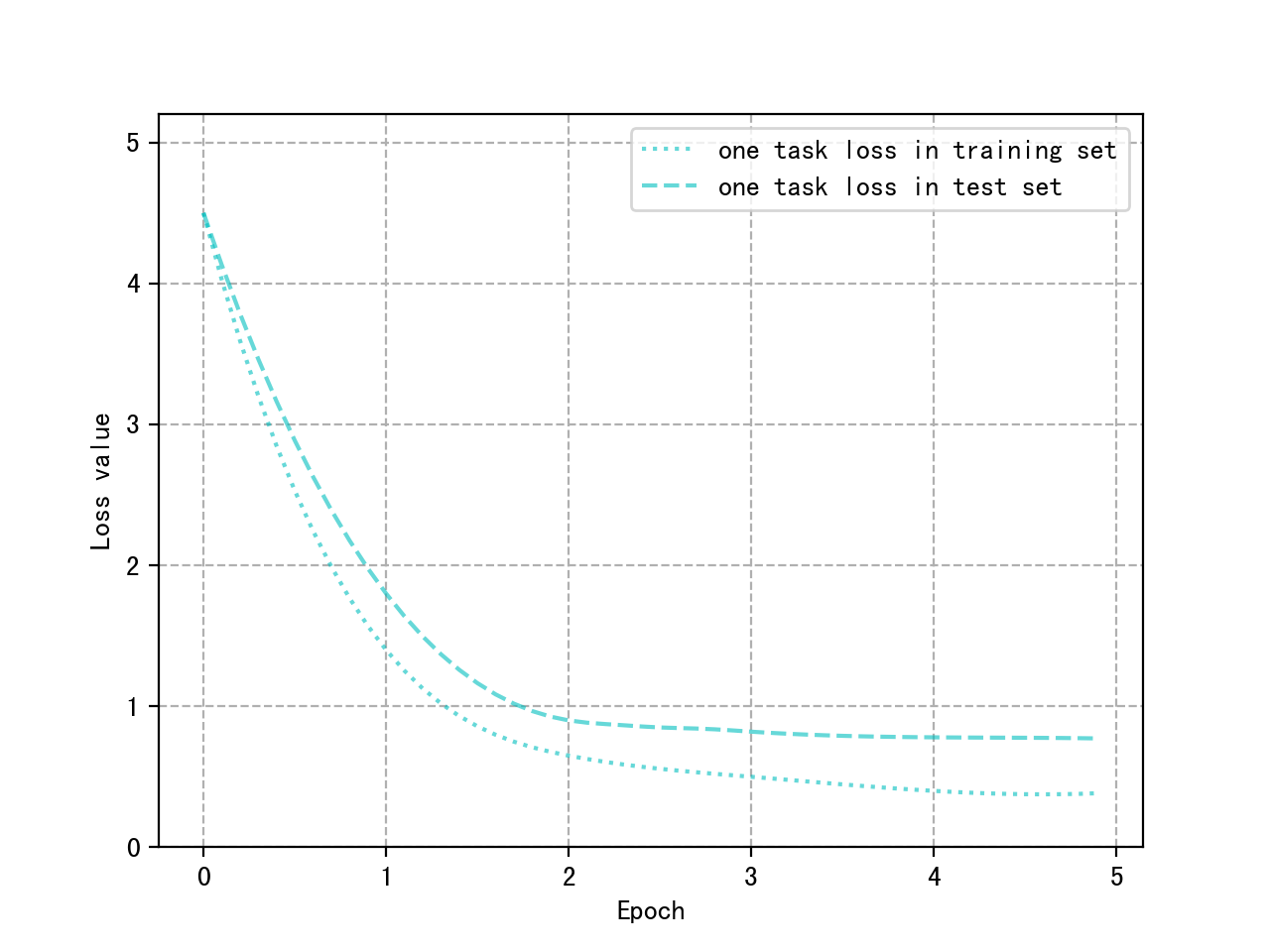}\label{loss_one_task}}

\subfigure[The loss-epoch diagram of the multi-task]{\includegraphics[scale=0.45]{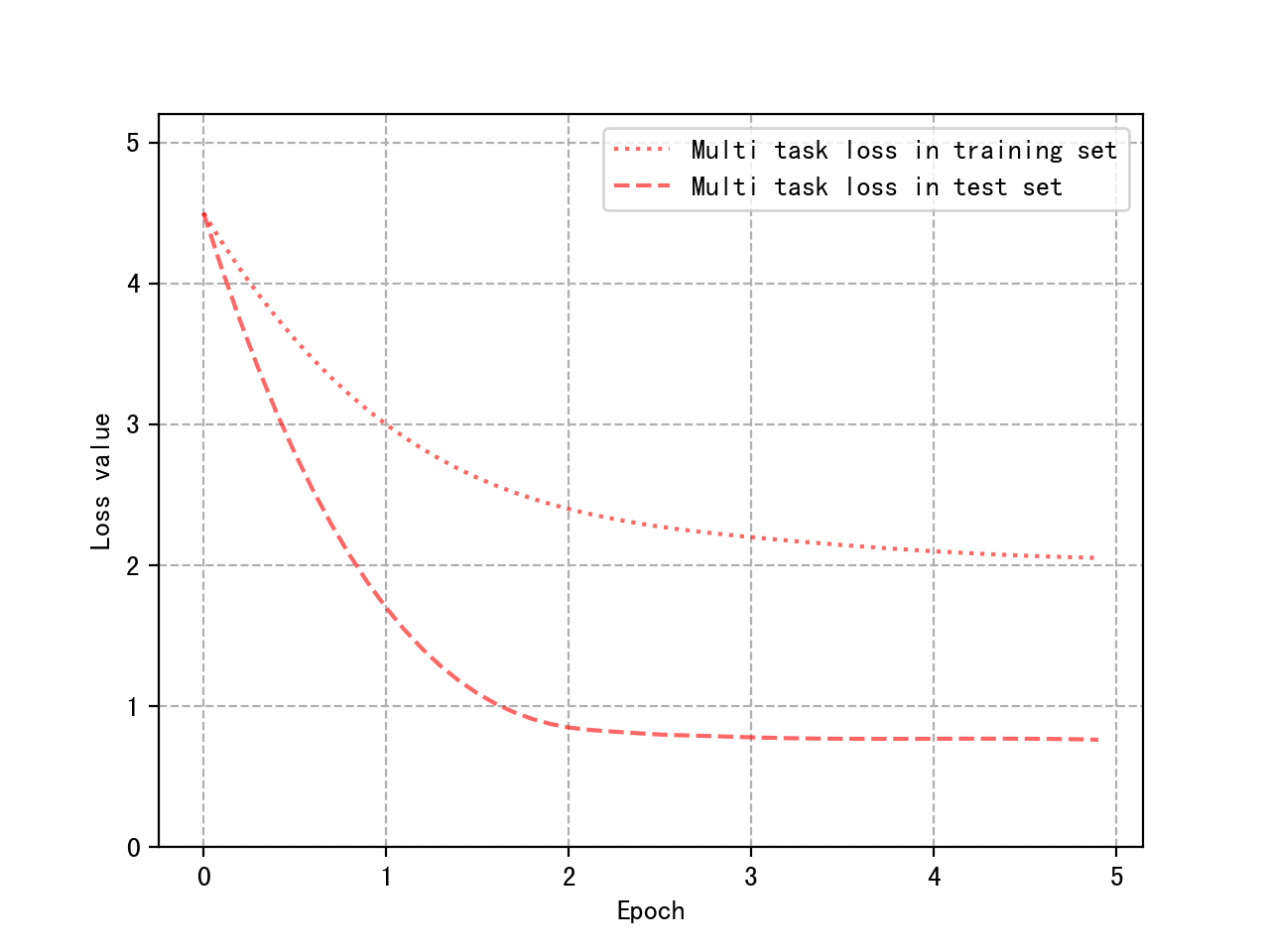}\label{loss_multi_task}}
\end{center}
\caption{The comparison of loss evolution.}
\label{loss_evolution}
\end{figure}

As shown in Fig.\ref{loss_evolution}, we compare the loss evolution of single and multi-task in both training and test dataset. For the single task, its training loss drops rapidly to a very low level and is always less than test loss. However, the results turn to be reversed for multi-task designed in this paper, the training loss always maintains a relatively high value and larger than the one for test.
These indicate that multi-task can indeed prevent the model from overfitting.

\section{The Visualization of Experts' Weights for Different Videos}

\begin{figure*}[bh]
\begin{center}
\includegraphics[scale=0.2]{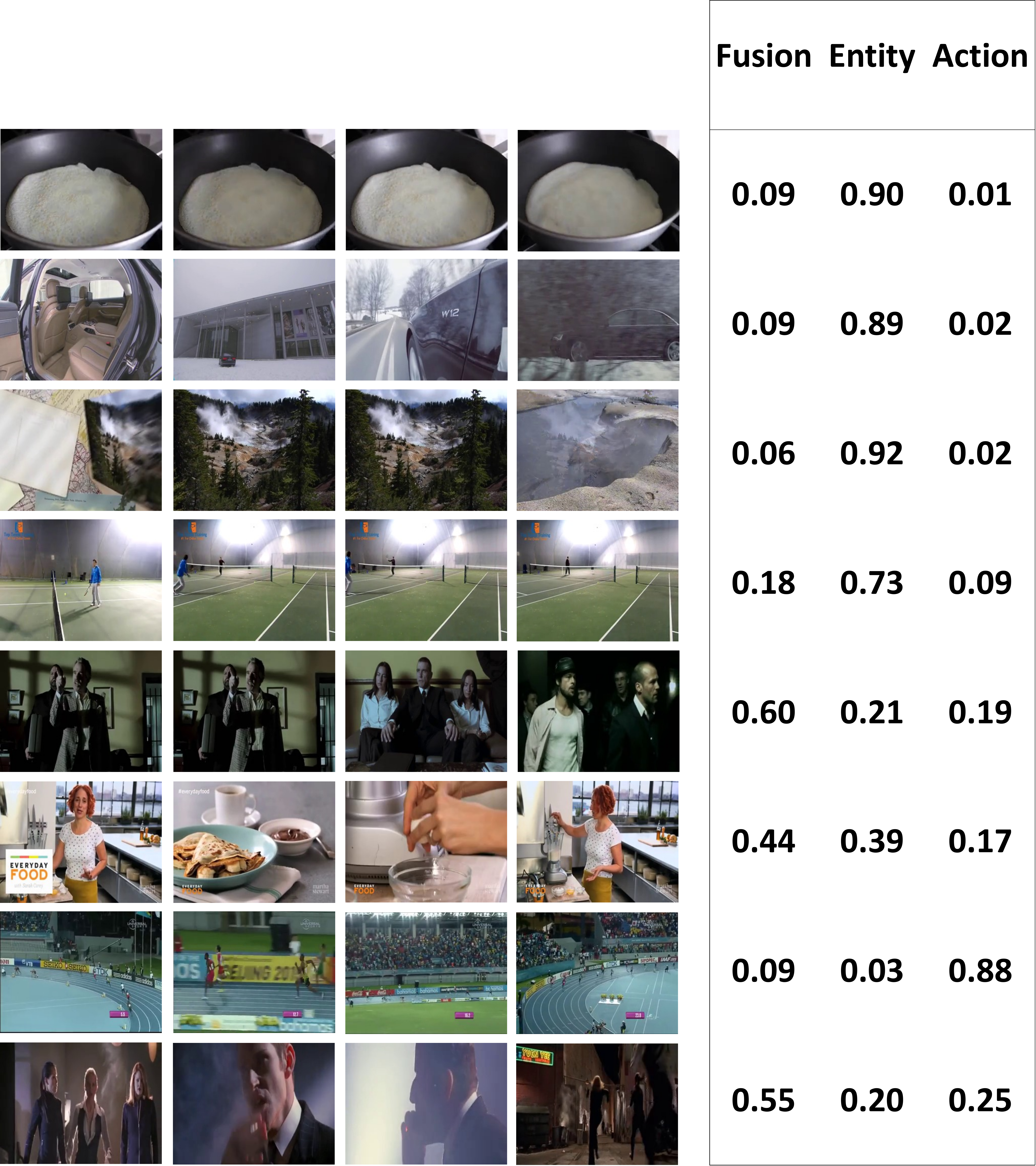}\caption{The visualization of experts' weights for different videos}
\label{bijiao_expert_weight}
\end{center}
\end{figure*}
To further verify whether the gating module has learned to assign appropriate weights for diverse videos and whether the extra two experts work, the weights for videos are tested as shown in Fig.\ref{bijiao_expert_weight}. The weights projected from the gate do vary a lot from video to video. From the first four exhibitions, where entity expert plays a crucial role, we can conclude that entity expert is relatively more critical when entity objects are more conspicuous, which means:
 \begin{itemize}
     \item The contents, styles, and perspectives of the frames are fairly consistent.
     \item There may exist homogeneous entities, which cannot exhibit various behaviors.
 \end{itemize}

Referring to the last four exhibitions, though action expert still occupies a lower weight, the results are much more larger than the average value which is 0.08. We have explained that the lower action expert weight may be caused by CLIP pre-training, and may conclude that action expert imposes an impact for the following two situations:
\begin{itemize}
    \item  There are apparent entities in the video that can perform specific behaviors. They are most likely people.
    \item The overall contents and perspectives are quite different, and it may be tough to distinguish specific actions from one frame.
\end{itemize}

\section{Further Experiments}
We enforced the experiments on other Three datasets:
\begin{itemize}
    \item \textbf{MSR-VTT full}, different from the 1k-A split reported in the text body, splits the whole dataset as 7k for training and 3k for test. And all the captions of test data are token into consideration.
    \item \textbf{DiDeMo}\cite{anne2017localizing} contains about 10000 videos range from 12s to 429s. We conform the tradition to concatenate all captions of a video into its text query as previous works \cite{luo2021clip4clip,Liu2019a}.
    \item \textbf{Activitynet}\cite{caba2015activitynet} is comprised of 20000 videos, whose descriptions are concatenated into one query. The corresponding splits are the same as \cite{luo2021clip4clip,Liu2019a}.
\end{itemize}

\begin{table}[h]
\centering
\caption{The experiments on MSR-VTT full, DiDeMo, and Activitynet.}
\label{extra_experiment}
\begin{tabular}{c|c|c|c|c}
\hline
&\multicolumn{2}{c|}{Text-to-Video} & \multicolumn{2}{c}{Video-to-Text} \\
\hline
Dataset &R@1  &R@5 &R@1 &R@5 \\
\hline
MSR-VTT full &48.8 &75.6 &50.3 &74.6 \\
DiDeMo &43.8 &71.4 &45.5 &71.2 \\
Activitynet &51.0 &77.7 &49.9 &77.4 \\
\hline
\end{tabular}
\end{table}
The results show that the proposed method is of great generalization and can achieve SOTA in various datasets.

\end{document}